\documentclass[conference]{IEEEtran}
\IEEEoverridecommandlockouts
\usepackage{cite}
\usepackage{amsmath,amssymb,amsfonts}
\usepackage{algorithmic}
\usepackage{graphicx}
\usepackage{textcomp}
\usepackage{xcolor}
\usepackage{subcaption}
\usepackage{dsfont}
\usepackage{multirow}
\usepackage[ruled,vlined]{algorithm2e}
\usepackage{enumitem}

\usepackage{hyperref}
\usepackage[normalem]{ulem}

\newcommand\blfootnote[1]{%
  \begingroup
  \renewcommand\thefootnote{}\footnote{#1}%
  \addtocounter{footnote}{-1}%
  \endgroup
}

\usepackage{cleveref}

\def\BibTeX{{\rm B\kern-.05em{\sc i\kern-.025em b}\kern-.08em
    T\kern-.1667em\lower.7ex\hbox{E}\kern-.125emX}}
\begin{document}
\bstctlcite{IEEEexample:BSTcontrol}
\title{Open-World Class Discovery with Kernel Networks\\
}

\author{\IEEEauthorblockN{Zifeng Wang\textsuperscript{*}, Batool Salehi\textsuperscript{*}, Andrey Gritsenko\textsuperscript{*}, Kaushik Chowdhury\textsuperscript{*}, Stratis Ioannidis\textsuperscript{*}, Jennifer Dy\textsuperscript{*}}
\IEEEauthorblockA{\textit{Department of Electrical and Computer Engineering} \\
\textit{Northeastern University}\\
Boston, MA \\
\textsuperscript{*}\{zifengwang, bsalehihikouei, agritsenko, krc, ioannidis, jdy\}@ece.neu.edu}
}

\maketitle


\begin{abstract}
\blfootnote{Accepted to the IEEE International Conference on Data Mining 2020 (ICDM'20)}
We study an \emph{Open-World Class Discovery} problem in which,  given labeled training samples from \emph{old} classes, we need to \emph{discover} \emph{new} classes from unlabeled test samples. 
There are two critical challenges to addressing this paradigm: (a) transferring knowledge from old to new classes, and (b) incorporating knowledge learned from new classes back to the original model. 
We propose \emph{Class Discovery Kernel Network with Expansion} (CD-KNet-Exp), a deep learning framework, which utilizes the Hilbert Schmidt Independence Criterion to bridge supervised and unsupervised information together in a systematic way, such that the learned knowledge from old classes is distilled appropriately for discovering new classes. 
Compared to competing methods, CD-KNet-Exp shows superior performance on three publicly available benchmark datasets and a challenging real-world radio frequency fingerprinting dataset.
\end{abstract}


\section{Introduction}
In the conventional supervised learning setting, we assume that we know all classes in advance; i.e., the classes that appear in the test set will be a subset of classes in the training set. This has been termed as the \emph{closed-world} assumption \cite{bendale2015towards, scheirer2012toward}; recent advances in deep learning \cite{lecun2015deep} have given impressive performance on supervised learning problems where the closed-world assumption holds, such as computer vision \cite{he2016deep, krizhevsky2012imagenet} and natural language processing \cite{DBLP:conf/naacl/DevlinCLT19}. However, in real world applications, we often encounter an \emph{open-world} \cite{scheirer2012toward, bendale2016towards, fei2016breaking} setting, in which unlabeled test samples come from new, previously unseen classes. This would be the case when a trained classifier is deployed in a completely new environment. 

As a concrete example, consider a  classifier that has been trained to distinguish between different breeds of dogs. An  open-world class discovery problem  would arise if we had access to this  trained classifier, and we ask to classify a wholly different test set, containing, e.g., different breeds of cats. Ideally, we would like to somehow leverage the dog classifier to distinguish between cat breeds. Though clearly, we cannot name cat breeds in this setting, it is possible that,  by incorporating the knowledge learned from dogs, we would be able to \emph{discover} the new cat breeds, clustering samples from the same breeds together. 

This problem constitutes the \emph{open-world class discovery problem} \cite{shu2018unseen, nixon2020semi}. 
%
Open-world class discovery poses a significant challenge, as transferring learned knowledge on old classes to new classes is not straightforward. Machine learning models may overfit to old classes; as a result,  knowledge learned, particularly through latent representations, may not generalize well to new classes. The more dissimilar old and new classes are, the more pronounced this problem becomes. Identifying which knowledge to transfer and leverage from old classes when trying to discover new classes is not trivial. This is further exacerbated in the case of deep models, that are by nature less interpretable. Coming up with an automated, principled way of extracting commonalities among representations is the main obstacle behind open-world settings. 


We address these challenges via an algorithm for leveraging deep architectures to solve the open-world class discovery problem. We first train a classifier on known classes.  We subsequently retrain it in the presence of unlabeled samples of new, previously unseen classes using an objective based on the Hilbert Schmidt Independence Criterion (HSIC). Intuitively, our scheme fine-tunes the latent representation obtained over the old-class dataset so that, when used to map new-class samples, the resulting images span a low rank subspace and are (jointly) well-clusterable.
This is accomplished even for deep models, whose latent representation is difficult to interpret.

Formally, our contributions are as follows:
\begin{itemize}
    \item We propose a deep learning framework which utilizes HSIC to bridge supervised and unsupervised information together in a systematic way, so that  learned knowledge from old classes is distilled appropriately for new classes. Our approach addresses  overfitting to old classes,  leading to improved class discovery, and can be generically applied to a broad array of deep architectures.
     \item Our algorithm, CD-KNet-Exp, shows superior performance on three benchmark datasets, including MNIST, Fashion-MNIST, CIFAR-100, and also a real world radio frequency fingerprinting dataset. In particular, it outperforms competitors over three benchmark datasets by a $2\%$-$12\%$ margin. 
\end{itemize}
The rest of this paper is organized as follows. We present related work in Section \ref{Sec:related workd}. Next, in Section \ref{Sec:Open world class discovery}, we formally define the open world class discovery problem and introduce our notation. We give an overview of  HSIC and its uses for supervised and unsupervised learning in Section \ref{sec:hsic}. In Section \ref{sec:methodology}, we present our three-stage framework  in detail. Our experimental results are provided in Section \ref{Sec:results and disscusions}; finally, we conclude in Section~\ref{Sec:conclusions}. 



\section{Related Work}
\label{Sec:related workd}

\noindent\textbf{Novelty Detection.}
A large body of prior work has been focused on novelty detection \cite{pimentel2014review,chen2016lifelong}, where the task is to design a model capable of both classifying instances that belong to the known training classes, and detect instances belonging to novel emerging classes at the same time. This differs from our setting, as the goal is to detect from a mixture of old and new samples which ones are old and which ones are new, without clustering the latter. Recently, a number of novelty detection methods were proposed based on kernel density estimation \cite{subramaniam2006online, bengio2006non}, nearest neighbor \cite{angiulli2002fast, hautamaki2004outlier, zhang2006detecting} and recent advances in deep learning \cite{chalapathy2018anomaly, gritsenko2019finding}. Our setting is orthogonal/complementary: once novel samples are detected, our method can be applied to discover new classes. 

\noindent\textbf{Semi-Supervised Learning.} Our problem setting seems close to semi-supervised learning \cite{grira2004unsupervised, zhu2009introduction}, where some samples are labeled and others are not.
However, in semi-supervised classification problem, \textit{all classes are known}
and every class has a corresponding labeled portion:  some of the samples for each class are labeled and the others are unlabeled. Information transfer can be achieved via deep learning models with great representational power \cite{DBLP:conf/iclr/KipfW17, rasmus2015semi}. 
Additional information provided on the samples can be leveraged \cite{yin2010semi, basu2004probabilistic}, such as must-link and cannot-link constraints. 
In contrast, our task aims to discover unseen new classes with no direct constraint information about these unseen new classes.  All the knowledge we have comes from the labeled old classes. Hence, in the open-world class discovery problem, figuring out how to transfer knowledge learned from  old classes to  new classes is a critical challenge that needs to be addressed.

\noindent\textbf{Open-World Class Discovery.}
To the best of our knowledge, very limited research has been performed in the area of class discovery. 
Recently, Nixon et al. \cite{nixon2020semi} proposed to train a neural network classifier on  old classes, followed by applying the K-means \cite{lloyd1982least} algorithm to directly cluster the new classes on the features extracted by the trained network. They provide two strategies for 
adding the new discovered classes back to the classifier:
static, where all new classes are added at once; 
and dynamic, where a single, most appropriate, class is added. 
Shu et al. \cite{shu2018unseen} use a pairwise network to learn a proper distance metric from seen old classes and utilize that metric for clustering the unsupervised data to discover new classes. 
Nixon et al. \cite{nixon2020semi} use a feature extractor that is trained only on the seen old classes. Similarly, Shu et al. \cite{shu2018unseen} train their distance metric only on the seen old classes.
Training only on the old classes may not be appropriate for the new classes.  CD-KNet, on the other hand, discovers new classes by learning a feature extractor that leverages information from both supervised seen old classes and the unsupervised data.

\section{Open World Class Discovery}
\label{Sec:Open world class discovery}

In this section, we provide a precise formulation of the open world class discovery problem; Table~\ref{tab:notation} summarizes our notation.
First, we are given a labeled dataset $\mathcal{D}_l = \{({x}_{i}, y_{i})\}_{i=1}^n$, where ${x}_i \in \mathbb{R}^{d_0}$ is the input sample and $y_i \in \mathcal{L}$ is the class label, from $m_l=|\mathcal{L}|$ classes. 
We  are also given an unlabeled dataset $\mathcal{D}_u = \{{x}_{j}\}_{j=1}^{n'}$, where ${x}_{j} \in \mathbb{R}^{d_0}$. These unlabeled samples belong to wholly distinct \textit{new classes}, that are not present in $\mathcal{D}_l$. That is, each sample $x_j\in \mathcal{D}_u$ is associated with class label $y_j\in \mathcal{L}'$, where $\mathcal{L}'$ is again a finite set of size $m_u=|\mathcal{L}'|$ such that $\mathcal{L}\cap \mathcal{L}'=\emptyset$. 
Our goal is to (a) train a classifier on 
 $\mathcal{D}_l$, and (b) leverage it over $\mathcal{D}_u$, so that we can discover (latent) ground truth classes in $\mathcal{D}_u$. Of course, $\mathcal{L}'$ cannot be discovered per se. Our objective is to therefore more precisely stated as clustering groups of $x_j\in \mathcal{D}_u$ if they share the same (unseen) label in $\mathcal{L}'$.
We assume that the number of new classes $m_u$ is known. 

Note that we have made two assumptions: (a) unlabeled samples only come from new classes and (b) the number of new classes are known.  We can relax the first assumption, for example, by applying a novelty detector first to filter out old classes (see, e.g., \cite{ pimentel2014review, markou2003novelty}).   Moreover, we can discover $m_u$ through standard methods  from clustering literature (see, e.g.,  \cite{lughofer2012dynamic}). 



\begin{table}[!t]
    \centering
    \caption{Notation Summary}
    \vspace{-2mm}
    \begin{tabular}{|c|l|}
    \hline
        Notation & Description \\
        \hline
        $\mathcal{D}_l$ & A labeled dataset \\
        $x_i$ & Input sample $i$ from $\mathcal{D}_l$ \\
        $y_i$ & Corresponding label of sample $i$ from $\mathcal{D}_l$ \\
        $n$ & Number of samples in the labeled dataset $\mathcal{D}_l$\\
        $\mathcal{L}$ & Set of known labels \\
        $m_l$ & Cardinality of set $\mathcal{L}$ (number of old classes) \\
        $\mathcal{D}_u$ & An unlabeled dataset \\
        $x_j$ & Input sample $j$ from $\mathcal{D}_u$ \\
        ${n}'$ & Number of samples in the unlabeled dataset $\mathcal{D}_u$\\
        $\mathcal{L}'$ & Set of new classes  \\
        $m_u$ & Cardinality of set $\mathcal{L'}$ ( number of new classes) \\
        $d_{0}$ & Dimension of input samples $x_i$, $x_j$\\
        $K_P$ & Kernel matrix of input samples matrix $P$ \\
        $K_Q$ &  Kernel matrix of labels matrix $Q$\\
        $\mathbb{H}(P, Q)$ & Hilbert Schmidt Independence Criterion between two \\
         & datasets, $P$ and $Q$ -- Eq.~(\ref{eq:hsic}) \\
         $\text{tr}(\cdot)$ & Trace of a matrix \\
        $\theta$ & Parameter of feature extractor\\
        $f_\theta(X)$ & Latent feature embedding of data $X$ parametrized by $\theta$ \\
        $d$ & Reduced dimension after applying feature extractor\\
        $X_l$ & Data matrix of labeled dataset $\mathcal{D}_l$\\
        $X_u$ & Data matrix of unlabeled dataset $\mathcal{D}_u$\\
        $Y$ & Label matrix of labeled dataset $\mathcal{D}_l$\\
        $X$ & Data matrix of both labeled and unlabeled datasets \\
        $U$ & latent emebeding of $X$ \\
        $\mathbb{H}_o$ & CD-KNet objective (Eq.~\eqref{eq:CD-KNet})\\
        $\lambda$ & Control parameter in Eq.~\eqref{eq:CD-KNet}\\
        $\hat{y}_j$ & Pseudo-labels/cluster assignments for ${n}'$  samples\\
        $\gamma$ & Learning rate\\
        $D$ & Degree matrix (Eq.~\eqref{eq:degree_matrix})\\
        $n_1$ & Number of samples from $\mathcal{D}_l$ used in computing $\mathbb{H}_o$\\
        $n_1'$ & Number of samples from $\mathcal{D}_u$ used in computing $\mathbb{H}_o$\\
        \hline
    \end{tabular}
    \label{tab:notation}
    \vspace{-5mm}
\end{table}

\section{Hilbert Schmidt Independence Criterion}
\label{sec:hsic}
The Hilbert Schmidt Independence Criterion (HSIC) \cite{gretton2005measuring} is a statistical dependence measure between two random variables. Just like Mutual Information (MI), it  captures non-linear dependencies between the random variables. Compared to MI, its  empirical computation is easy, avoiding the explicit estimation of joint probability distributions. On account of this, it has been widely applied in different domains, such as feature selection \cite{Song2012}, dimensionality reduction \cite{niu2011dimensionality}, alternative clustering  \cite{wu2018iterative}, and deep clustering \cite{wu2020deep}.

Formally, consider a set of i.i.d. sample tuples $\{(p_i, q_i)\}_{i=1}^{N}$, where $p_i \in \mathbb{R}^d$, $q_i \in \mathbb{R}^c$. Let $P \in \mathbb{R}^{N \times d}$ and $Q \in \mathbb{R}^{N \times c}$ be the matrices whose rows are the corresponding samples. Also, let $k_p : \mathbb{R}^d \times \mathbb{R}^d \rightarrow \mathbb{R}$ and $k_q : \mathbb{R}^c \times \mathbb{R}^c \rightarrow \mathbb{R}$ be two characteristic kernel functions for $p_i$ and $q_i$, respectively. Examples are  \begin{align}
    k_P(p_i, p_j) = e^{-\frac{\Vert p_i - p_j, \Vert^2}{2\sigma^2}}
    \end{align}
    i.e., the Gaussian kernel, and the linear kernel 
    \begin{align}
    k_Q(q_i, q_j) = q_i^\top q_j.
    \end{align}
     We further define $K_P, K_Q $ to be the kernel matrices for $P$ and $Q$ respectively, where $K_{P} = \{k_P(p_i, p_j)\}_{i,j}\in \mathbb{R}^{N\times N}$ and $K_{Q} = \{k_Q(q_i, q_j)\}_{i,j}\in \mathbb{R}^{N \times N}$.

The HSIC between $P$ and $Q$ is estimated empirically with kernels $k_P,k_Q$ via:
\begin{align}
    \label{eq:hsic}
    \mathbb{H}(P, Q) = \frac{1}{(N-1)^2}\text{tr}(K_PHK_QH),
\end{align}
where $H_{i,j}=\delta_{i,j} - N^{-1}$. Intuitively, HSIC measures the dependence between  the  random variables $p,q$ from which the i.i.d.~samples $\{(p_i, q_i)\}_{i=1}^{N}$ where generated.

\subsection{Supervised learning setting}
\label{sec:supervisedHSIC}
Consider a data matrix $X \in \mathbb{R}^{N\times d_0}$, containing $N$ $d_0$-dimensional samples per row,  and label matrix  $Y \in \{0,1\}^{N \times m}$, representing the one-hot encoding of $m$ labels.
We can utilize HSIC to perform dimensionality reduction in this supervised learning setting \cite{wu2019solving}. We can do so by maximizing the dependency between a non-linear feature mapping of input $X$ and labels $Y$ as follows. Let  $f : \mathbb{R}^{d_0} \mapsto \mathbb{R}^{d}$, where $d\ll d_0$, be a feature extractor, e.g. neural network, parameterized by $\theta$. Denote by $f(X)\in \mathbb{R}^{N\times d}$ the matrix of images of  rows (i.e., samples) in $X$.
We can substitute $f_\theta(X), Y$ for $P, Q$ in Eq.~\eqref{eq:hsic},   We set $K_X$ as a Gaussian kernel and $K_Y$ as a linear kernel. Then the solution of the following optimization problem:  
\begin{align}
    \underset{\theta}{\operatorname{max}} &\quad \mathbb{H}(f_\theta(X), Y),
\end{align}
maximizes the dependence of $f_\theta(X)$ and $Y$. Intuitively, this forces the feature extractor to be maximally correlated with $Y$.  Having reduced dimensions thusly, a shallow classifier (e.g., logistic regression) can be used to learn the labels from the lower dimensional images $f_\theta(X)$.

\subsection{Unsupervised learning setting}
\label{sec:unsupervisedHSIC}
In the unsupervised case, we are only given data matrix $X \in \mathbb{R}^{N\times d_0}$. We can utilize HSIC to perform unsupervised learning by maximizing the dependency between a non-linear feature mapping of input $X$ and a \emph{learnable} latent cluster embedding matrix $U \in \mathbb{R}^{N \times c}$ (see \cite{wu2020deep,niu2011dimensionality}). We substitute $P, Q$ with $X, U$. Under the unsupervised setting, we set the kernel matrix for $X$ as a \emph{normalized} Gaussian kernel:
\begin{align}
    \tilde{K}_X = D^{-\frac{1}{2}} K_X D^{-\frac{1}{2}},
\end{align}
where $D$ is the degree matrix defined by:
\begin{align}
    \label{eq:degree_matrix}
    D=\operatorname{diag}\left(K_{X} \mathbf{1}_{N}\right) \in \mathbb{R}^{N \times N}.
\end{align}
We also use linear kernel $K_U=UU^\top$ for $U$.
Consider the following optimization problem:
\begin{subequations}
\label{eq:spclustering}
    \begin{align}
    \underset{U}{\operatorname{max}} &\quad \mathbb{H}(X, U), \\
    \text{s.t.} &\quad U^\top U = I.
    \end{align}
\end{subequations}
The optimal solution $U_0$ is the spectral clustering embedding of $X$ (see \cite{niu2011dimensionality} for a proof). Thus, HSIC provides an alternative perspective to perform spectral clustering.

Moreover, analogous to the supervised setting, when feature extractor $f_\theta$ is introduced, we can  joint optimize $\theta, U$ via :
\begin{subequations}%
\label{eq:deepspclustering}
    \begin{align}
    \underset{\theta,U}{\operatorname{max}} &\quad \mathbb{H}(f_\theta(X), U), \\
    \text{s.t.} &\quad U^\top U = I.
    \end{align}
\end{subequations}
HSIC enforces the feature extractor to learn a non-linear mapping $f_\theta(X)$ of input $X$ to match to spectral clustering embedding $U$~\cite{wu2020deep}. 


\section{Proposed Class Discovery \\Kernel Network Approach}
\label{sec:methodology}

In this section, we provide an overview of our proposed approach, describe Class Discovery Kernel Network (CD-KNet) for solving the open-world class discovery problem, and present  a neural network expansion scheme that introduces  information feedback from (discovered) new classes. 
\begin{figure}
    \centering
    \includegraphics[width=\linewidth]{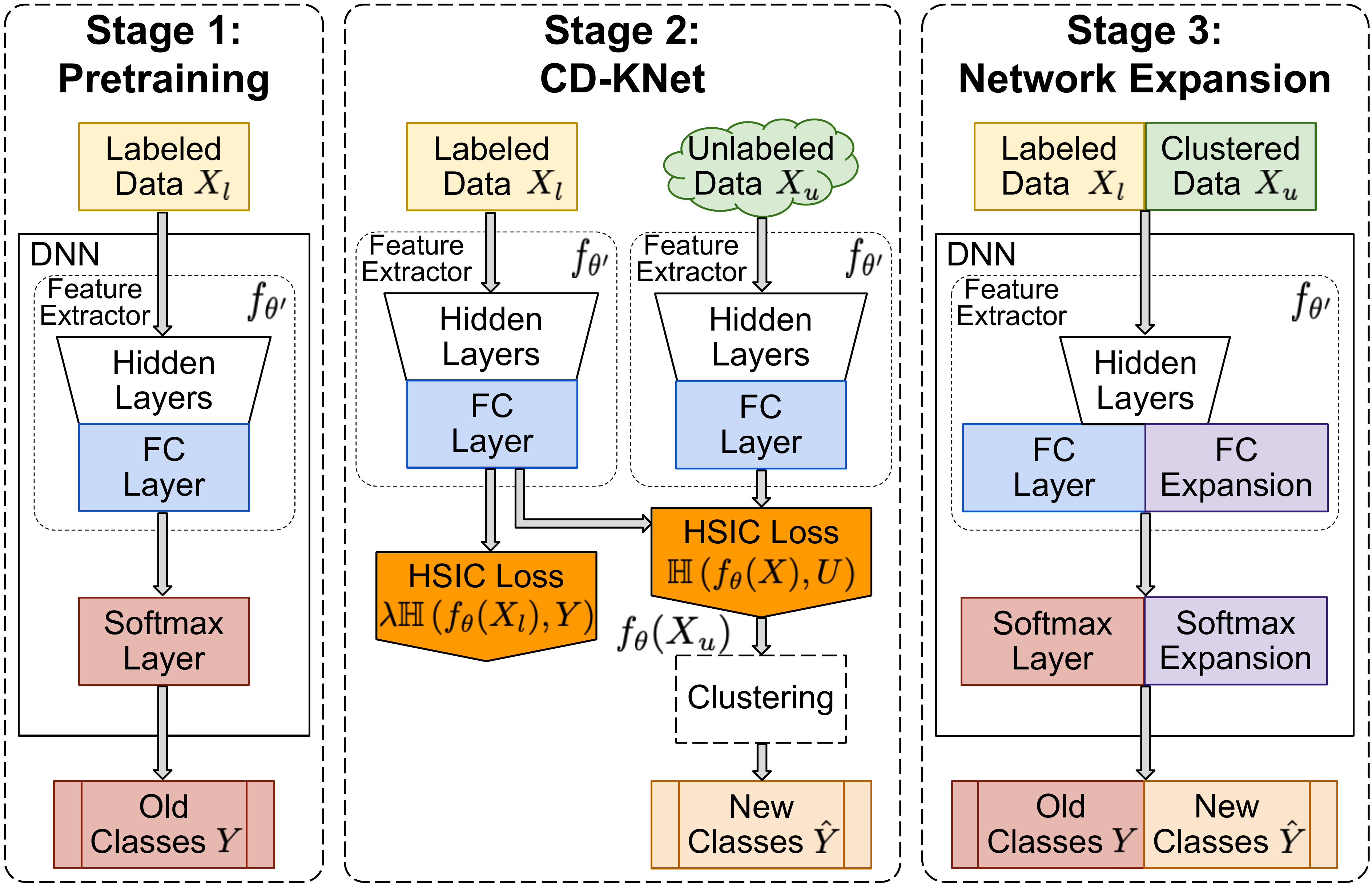}
    \caption{Overview of the CD-KNet-Exp algorithm.
    In Stage 1, we train a classifier (deep convolutional neural network) on labeled data $X_l$ from seen old classes. In Stage 2, we fine tune the feature extractor by leveraging both labeled data $X_l$ and unlabeled data $X_u$ through two HSIC-based penalties. We then cluster the unlabeled data in the learned feature embedding $f_\theta(X_u)$ to generate pseudo-labels $\hat{Y}$. Finally, in Stage 3, the classifier's penultimate (feature embedding) and last (softmax classification) layers are expanded to adjust the model to learn from both the old and new classes. }
    \label{fig:network_overview}
    \vspace{-5mm}
\end{figure}

\subsection{An Overview of CD-KNet with Expansion}
\label{subsection:overview}

In this paper, we propose Class Discovery Kernel Network (CD-KNet) and  Class Discovery Kernel Network with Expansion (CD-KNet-Exp). 
Our framework breaks the open-world class discovery problem into three stages, shown in   \Cref{fig:network_overview}. We describe them in detail below.

During stage one, we train a deep neural network (DNN)~\cite{lecun2015deep} classifier from the labeled dataset $\mathcal{D}_l$. 
One can view the DNN as a combination of a non-linear feature extractor followed by a softmax classifier: the first layer until the penultimate layer of a DNN constitutes the feature extractor. Assuming the penultimate layer has $d$ dimensions (neurons), we denote this feature extractor as $f_\theta:\mathbb{R}^{d_0}\to \mathbb{R}^d$, parameterized by $\theta$. As usual, we refer to the output of this feature extractor $f_\theta(\cdot)\in\mathbb{R}^d$ as the \textit{latent embedding}.

In stage two, we use the unsupervised data $\mathcal{D}_u$.  Our goal is to learn from both the labeled dataset $D_l$ and the unlabeled data $\mathcal{D}_u$ for discovering the new (unseen) classes.  To that end, we fine tune our feature extractor $f_\theta$ through our CD-KNet algorithm described in the next section. 
CD-KNet bridges both the supervised \emph{old classes} and the unsupervised cluster (\emph{new class}) discovery task through a use of two HSIC-based penalties. 
In addition to learning an updated latent embedding $f_\theta (X)$, CD-KNet also learns cluster assignments to data samples from $\mathcal{D}_u$.  We refer to these new labels as \textit{pseudo-labels}, as they abstract the newly discovered classes.

In stage three, we expand our deep network, CD-KNet-Exp, by expanding the original network with additional latent embedding nodes and output nodes to learn from both the \textit{old} and \textit{new} classes utilizing labels in $D_l$ and the \textit{pseudo-labels} in $D_u$.
This further fine-tunes both the network and our classification outcomes; it is also attuned to (and exploits) the linear separability of the latent embeddings learned by our feature extractor. Thus, whenever a new test sample comes, as long as it belongs to the $(m_l + m_u)$ classes, CD-KNet-Exp can provide the prediction via the expanded neural network.

We elaborate on the details of CD-KNet and CD-KNet-Exp in the following subsections.

\subsection{Stage 1: Pre-training the model}

As discussed above, pre-training the model involves training feature extractor $f_\theta:\mathbb{R}^{d_0}\to \mathbb{R}^d$, as well as a final dense/softmax layer. This can be trained  over $\mathcal{D}_l$ via classic methods, e.g., via stochastic gradient descent (SGD) over standard loss functions (square loss, cross-entropy, etc.).

\subsection{Stage 2: Class Discovery Kernel Network (CD-KNet)}
At the conclusion of stage one, we have learned the feature extractor $f_\theta$ from the DNN. 
In stage two, our goal is to discover \textit{new classes}.  
A simple solution would be to leverage the learned feature extractor $f_\theta$ to map $D_u$ to the learned feature embedding space and directly perform clustering in this space (see, e.g., \cite{nixon2020semi}). However, it is possible that the feature embedding space learned from the \textit{old classes} in $\mathcal{D}_l$ are highly biased to the \textit{old classes}. As a result, the learned embedding may not generalize well to $\mathcal{D}_u$. 
In our work, instead of using $f_\theta$ from stage one directly, we update it, forcing it to jointly adapt to  both the supervised old classes in $\mathcal{D}_l$ as well as the unsupervised new data in $\mathcal{D}_u$.



Formally, following the notations in Section~\ref{sec:hsic}, let $X_l \in \mathbb{R}^{n \times d_0}$, $Y \in \mathbb{R}^{n \times m_l}$ be the data and label matrix for the labeled dataset 
$\mathcal{D}_l$ respectively, where $Y$ comprises the one-hot encoding representation of labels. Similarly, let $X_u \in \mathbb{R}^{n' \times d_0}$ be the data matrix for the unlabeled dataset. We concatenate $X_l$ and $X_u$ to get $X \in \mathbb{R}^{(n+n') \times d_0}$, a matrix containing both labeled and unlabeled datasets. We also denote by $U \in \mathbb{R}^{(n+n') \times r}$, the corresponding latent embedding of $X$, where $r=m_l+m_u\in \mathbb{N}$ is the predefined dimensionality of the latent embedding. 
We learn the updated feature extractor $f_\theta$ and discover new classes by solving the following CD-KNet optimization problem:
\begin{subequations}
\label{eq:CD-KNet}
\begin{align}
    \underset{U, \theta}{\operatorname{max}}  &\quad \mathbb{H}_o (\theta,U)= \mathbb{H}(f_\theta(X), U) + \lambda \mathbb{H}(f_\theta(X_l), Y), \label{eq:ow-hsic}\\ 
    \text{s.t.} &\quad U^\top U = I,
\end{align}
\end{subequations}
where $\mathbb{H}$ is defined in Equation~(\ref{eq:hsic}) and $\lambda\geq 0$ is the control parameter between the supervised and unsupervised objectives. We describe how to solve Prob.~\eqref{eq:CD-KNet} below, in Section~\ref{subsec:solve_hsic}. Intuitively, the first term encourages the separation of all classes, old and new; both should be ``clusterable'', as captured by the high dependence with low-rank, orthogonal matrix $U$.  The second term introduces supervised information, ensuring that the latent embedding maintains the separation between old classes, as the latter remain aligned with their labels.

As a final step of the second stage, we take $f_\theta(X_u)$, the latent embedding of the new dataset, and cluster it. 
In more detail, upon convergence, the feature extractor $f_\theta$ has been refined to an extent that incorporates the information from labeled old classes as well as unlabeled new classes, resulting in a feature space which is able to separate both old and new classes well. In order to discover new classes, we feed all samples from $\mathcal{D}_u$ to the feature extractor to get $f_\theta(X_u)$, the matrix whose rows are latent feature embedding representations of samples from $\mathcal{D}_u$. We then can perform any clustering method, e.g. K-means, to get the cluster assignments $\{\hat{y}_{j}\}_{j=1}^{n'}\in \{1,\ldots,m_u\}^{n'}$. Note that these $m_u$ clusters constitute our  new classes. We refer to labels $\hat{y}_j$ as \emph{pseudo-labels}, as they correspond to our discovered classes (that are, ideally, consistent with the ground truth classes $\mathcal{L}'$).  

 The entire pipeline is illustrated in  Figure~\ref{fig:network_overview}, Stage 2. We call the pipeline as CD-KNet; the output of this stage, namely, the pseudo-labels, can be used as our final class discovery outcome. In practice, however, we further refine this with one additional stage, involving a network expansion. We describe this below. 
 Note that, via  Prob.~\eqref{eq:CD-KNet}, we leverage  supervised  information in two ways: first, via pre-training of the feature extractor, which is used as a starting point for the algorithm solving  Prob.~\eqref{eq:CD-KNet} below, as well as through enforcing the joint ``clusterability'' of both old and new latent embeddings.

\subsection{Stage 3: Network Expansion: CD-KNet-Exp}
\label{sec:extension}

In our final stage, once all new samples ${X}_{u}\in\mathcal{D}_u$ are assigned with pseudo-labels by CD-KNet, we 
use these labels to retrain the network, under an appropriate network expansion \cite{marnerides2018expandnet}. We describe this in detail here.
As mentioned earlier, a DNN can be regarded as the composition of a feature extractor $f_\theta$ and a softmax layer, i.e., the final dense layer with softmax activation. A simple heuristic to expand the network is just to expand the softmax layer by adding as many nodes as the number of new classes we have discovered. This strategy has been adopted in some prior works under a different context, such as transfer learning \cite{tan2018survey}. 

However,  we also need to consider is the \emph{representation capacity} of $f_\theta$. When old classes and new classes are combined, a feature extractor will naturally need more capacity, i.e. more parameters, to represent a more complex dataset. As suggested by Zeiler et al. \cite{zeiler2014visualizing}, shallower layers in DNN always extract general, abstract features which are common among different tasks, while deeper layers capture specific features closely related to the task/dataset. So we decide to only expand the final layer of the feature extractor, i.e. the penultimate layer of the whole DNN, and keep the rest of the feature extractor unchanged. In practice, we find that expanding shallower layers does not affect the final performance much as overfitting may happen easily at shallower layers.

To that end, in the third stage, we expand the network by adding $m_u$ to the last layer, and 25\% to the penultimate layer. The expanded model is then fine-tuned over both $\mathcal{D}_l$ and $\mathcal{D}_u$. In particular, the model is fine-tuned on $\mathcal{D}_u$ with pseudo-labels $\hat{Y}_u$ to incorporate new classes. 
In addition to learning from $\mathcal{D}_u$ and the new classes, we also include a fraction $p\%$ of the old classes $\mathcal{D}_l$ to strengthen previous learned knowledge from seen old classes. We refer to this as fine-tuning rather than training, because we lower the learning rate of the expanded model, except its expanded two final layers.
It is also possible to just retrain a new model, however, in our experiments, we find that fine-tuning the model always converges better and faster than  retraining a model from scratch.
Overall, the process of network expansion is summarized and presented in Stage~3 in~\Cref{fig:network_overview}. We refer to the complete 3-stage pipeline as CD-KNet-Exp. The final outputs of this process are the labels produced by the expanded model over $\mathcal{D}_u$.

\subsection{Solving the CD-KNet Optimization Problem}
\label{subsec:solve_hsic}

We adopt an alternating optimization strategy to learn $\theta$ and $U$ iteratively. The whole process is shown in Algorithm~\ref{alg:CD-KNet}.

\begin{algorithm}[!t]
\SetAlgoLined
\textbf{Input:} whole dataset matrix $X$, labeled dataset matrix $X_l$ and its corresponding label matrix $Y \in \mathbb{R}^{n \times m_l}$ \\
\textbf{Output:} parameter of feature extractor $\theta$, cluster assignments of $\mathcal{D}_u$\\
 \textbf{Initialization:} initialize $\theta$ by training on $\mathcal{D}_l$\; initialize $U$ by spectral embedding of $f_\theta(X)$\\
 \While{$\theta$ has not converged}{
  Update $\theta$ via one epoch of SGD via Eq. (\ref{eq:update1}) and (\ref{eq:update2}) alternatingly, while keeping $U$ fixed.\\
  Update $U$ via eigendecomposition according to Eq. (\ref{eq:update_D}). 
 }
 Cluster $f_\theta(X_u)$ to get cluster assignments.
 \caption{CD-KNet Algorithm}
 \label{alg:CD-KNet}
\end{algorithm}

\noindent\textbf{Initialization:} $\theta$ is initialized by the supervised training in stage one. We initialize $U$ by conducting spectral clustering on $f_\theta(X)$, which is equivalent to \eqref{eq:spclustering} \cite{niu2011dimensionality}.


\noindent\textbf{Updating $\theta$:} Assuming $U$ is fixed, we update $\theta$ via gradient ascent. In practice, we would like to update $\theta$ using stochastic gradient ascent via mini-batches. However, if we randomly sample mini-batches among all samples, we need to keep track of labeled and unlabeled parts inside each mini-batch. Thus, we simplify the process by optimizing the supervised and unsupervised part of $\mathbb{H}_o$ iteratively. First, we sample an unlabeled mini-batch $X_{b}$ from $X$, updating $\theta$ by:
\begin{align}
\label{eq:update1}
    \theta := \theta + \gamma \nabla \mathbb{H}(f_\theta(X_{b}), U_b),
\end{align}
where $\gamma$ is the learning rate and $U_b$ is the corresponding latent cluster embedding for this mini-batch. Then, we sample a mini-batch $X_{b}$ from $X_l$, updating $\theta$ by:
\begin{align}
\label{eq:update2}
    \theta := \theta + \lambda \gamma \nabla \mathbb{H}(f_\theta(X_{b}), Y_b),
\end{align}
where the learning rate $\gamma$ remains the same as in the previous step and $Y_b$ is the corresponding label matrix for this mini-batch. We update $\theta$ for an epoch before we update $U$.

\noindent\textbf{Updating $U$:} When $\theta$ is fixed, the optimization in Equation~(\ref{eq:CD-KNet}) w.r.t $U$ alone is equivalent to:
\begin{align}
\label{eq:CD-KNet-opt}
    \underset{U}{\operatorname{max}} &\quad \text{tr}(U^{\top}HD^{-1/2}K_{f_\theta(X)}D^{-1/2}HU), \\
    \text{s.t.} &\quad U^\top U = I,
\end{align}
where here we used $K_U = U U^\top$, $\text{tr}(\cdot)$ to represent the trace of a matrix and applied the cyclic property of the trace. This maximization problem of $U$ can be solved via eigendecomposition.
The optimal solution for $U$ is given by the top $r$ eigenvectors of the following matrix:
\begin{align}
    \label{eq:update_D}
    \mathcal{L}_{\theta}=H D^{-1 / 2} K_{f_{\theta}(X)} D^{-1 / 2} H,
\end{align}
where $H$ is as in Eq.~\eqref{eq:hsic}, and $D$ is the degree matrix given by Eq.~\eqref{eq:degree_matrix}.

\noindent\textbf{Subsampling:} While the number of labeled and unlabeled samples $n$, $n'$ could be very large, it is often time and resource consuming to compute $\mathbb{H}_o$ using all of the data. As an alternative, 
we perform subsampling to compute and optimize $\mathbb{H}_o$. Specifically, we sample $n_1$ labeled samples from $\mathcal{D}_l$ and $n'_1$ unlabeled samples from $\mathcal{D}_u$, where $n_1 \ll n$ and $n'_1 \ll n'$. We then form labeled data matrices $X_{l1} \in \mathbb{R}^{n_1\times d_0}$ as well as its corresponding label matrix $Y_1 \in \mathbb{R}^{n_1\times m_l}$, and unlabeled data matrix $X_{u1} \in \mathbb{R}^{n'_1 \times d_0}$. We concatenate $X_{l1}$ and $X_{u1}$ to get $X_1 \in \mathbb{R}^{(n_1 + n'_1)\times d_0}$ and define latent embedding $U_1 \in \mathbb{R}^{(n_1 + n'_1)\times r}$. Finally, we replace $X, U, X_l, Y$ with $X_1, U_1, X_{l1}, Y_1$ in Equation~(\ref{eq:CD-KNet}) and solve the corresponding optimization problem.
Our experiments show that using only 5\% of the original dataset  suffices to get a good performance.


\section{Experiments}
\label{sec:experiments}
In this section, we investigate how our proposed method CD-KNet-Exp compares against competing methods on
three benchmark datasets and one real world radio frequency fingerprinting dataset.  We also conduct comprehensive experiments to explore the effects of different controllable parameters on the performance of our algorithm.
\subsection{Datasets}
\label{sec:data}


Our proposed Class Discovery Kernel Network with Expansion (CD-KNet-Exp) method is evaluated on three benchmark datasets, \emph{MNIST}, \emph{Fashion MNIST} and \emph{CIFAR-100}, as well as a real world dataset of radio frequency transmissions (\emph{RF-50}). 

\noindent\textbf{MNIST.}
MNIST is a well-known database of grayscale images of handwritten decimal digits~\cite{lecun1998gradient}. The dataset 
contains $60,000$ digits in the training set, and $10,000$ digits in the test set. In our experiments, we select the first 5 digits as the \emph{labeled, old} classes, while the rest last 5 digits to constitute a set of \emph{unlabeled, new} classes. 

\noindent\textbf{Fashion MNIST.}
Fashion MNIST was first introduced by Xiao et al. in~\cite{xiao2017fmnist}, and contains grayscale images of 10 types of fashion products, including clothes, shoes, and accessories. Fashion MNIST follows the original MNIST dataset in image size and structure of training and test splits; however, each class is represented by exactly $7,000$ images. Again, a set of labeled classes consists of the first 5 fashion products, while the rest 5 classes are treated as unlabeled.

\noindent\textbf{CIFAR-100.}
CIFAR-100 is another imaging dataset that contains $60,000$ color images of 100 categories of objects, with $6,000$ images per category. Here, the dataset comprises of $50,000$ training and $10,000$ testing images. In analogy to MNIST datasets, object categories in CIFAR-100 are split into labeled and unlabeled, with the first 70 classes belonging to the former group and 30 included in the latter one.

\noindent\textbf{RF-50.}
This dataset contains $8,800$ radio transmissions from $50$ WiFi devices recorded in the wild~\cite{jian2020rfmls,riyaz2018deep}. 
Wireless signals undergo 
equalization 
\cite{riyaz2018deep}.
The dataset is split as follows: $141$ recordings from each device constitute the training subset, while $35$ are used for testing. Also, we randomly choose $35$ devices and mark them as labeled, i.e. they form $\mathcal{D}_l$ part of data, and the other $15$ devices are considered as unlabeled, i.e. form the $\mathcal{D}_u$ dataset.

\subsection{Competing Methods}
\label{sec:methods}

As mentioned in the related work section, open-world class discovery is a relatively new problem that is insufficiently investigated at the moment, and only one direct competitor with reproducible code is available~\cite{nixon2020semi}.
Nevertheless, we devise simple baselines via variations of our own framework. We also compare against  state-of-the-art deep clustering methods to strengthen our empirical results.

\noindent \emph{Our variants:} \\
\textbf{CD-KNet-Exp}. CD-KNet-Exp is our proposed class discovery kernel network approach as described in Section~\ref{sec:methods} and summarized in~\Cref{fig:network_overview}.
Next, we perform an ablation study and describe three modifications of our framework designed to understand the importance of each core component. 
\\
\textbf{CD-KNet}. To understand the importance of network expansion to accurately predict new classes $\hat{Y}$, we compare against CD-KNet which is a variant of our approach without the expansion (i.e., it only completes the first two stages of the framework). 
\\
\textbf{UCD-KNet-Exp}. Another crucial component of the proposed method is the incorporation of both \textit{supervised old class data} and the \textit{unlabeled data} in learning our feature extractor $f_\theta$ based on Objective~\ref{eq:ow-hsic}. 
To evaluate the influence of supervision, we set $\lambda = 0$ in Eq.~\ref{eq:ow-hsic}, and treat this unsupervised variant, UCD-KNet-Exp, as another competitor.
\\
\textbf{UCD-KNet}. Finally, we remove both network expansion and supervised components of the proposed CD-KNet-Exp framework, and consider this simplest unsupervised variant as UCD-KNet.

\noindent \emph{We compare against the current state-of-the-art approach to class discovery, Semi-Supervised Class Discovery (SSCD)~\cite{nixon2020semi}.} \\
\textbf{SSCD}.
Nixon et al.~\cite{nixon2020semi} propose a framework for a new class discovery based on the idea of (a) training a classifier on known classes, (b) applying it to unseen classes, (c) detecting new classes via K-means clustering, and (d) expanding the classifier via pseudo-labels. 
This is can be seen as a simple ``clustering plus supervised learning'' baseline compared to CD-KNet-Exp, skipping the additional retraining via HSIC.
Originally, the SSCD framework adopts the work of Hendrycs and Gimpel~\cite{DBLP:conf/iclr/HendrycksG17} to detect `new class'-candidate data samples. However, because novelty detection is not the focus of this paper, and in order to perform a fair comparison with our method, we assume that all test samples  do not belong to the known classes, eschewing the novelty detection component.
\\
\textbf{SSCD-Exp}. For their method, Nixon and others propose to expand a classification model solely in the last layer to accommodate the increased number of classes, and then retrain the model on data with both original and clustered labels.  Here, we adapt/modify the original SSCD method with our proposed strategy to expand their classification model. As in the 3rd stage of our method, we expand their neural network \emph{both in the penultimate and ultimate layers.} Hereinafter, we refer to this method as SSCD-Exp. 
Both SSCD and SSCD-Exp are implemented in accordance with the description and parameter settings provided in the original paper. We did not compare to Shu et al. \cite{shu2018unseen}, as no code is publicly available; we note however that the NMI they report on MNIST is 0.48, far lower than CD-KNet-Exp (0.856) and other competitors. 

\noindent \emph{Finally, we also compare against two state-of-the-art deep learning-based clustering algorithms.}\\
\textbf{Deep Embedding Clustering (DEC).} In~\cite{xie2016unsupervised}, the authors use stacked autoencoder to map the input $X_u$ to the low-dimensional feature space and then perform K-means clustering to initialize $m_u$ cluster centroids. The main idea behind DEC is to obtain probabilistic cluster assignments and then iteratively update them using the Kullback-Leibler divergence between the distribution of such ``soft'' assignment values and some auxiliary distribution. \\
\textbf{Deep Adaptive Clustering (DAC).} DAC~\cite{chang2017deep}
re-casts a clustering task to a binary pair-wise classification problem. DAC is an iterative method for assigning a pair of inputs to the same class if their embeddings are similar enough, assign to different classes if embeddings are different enough; otherwise, it discards the pair from the current training iteration. The similarity and dissimilarity thresholds in DAC are changed adaptively after each iteration.
In the experiments, we utilize the original implementations provided by the authors for both DEC and DAC methods.



\begin{figure}[tb]
    \includegraphics[width=\linewidth]{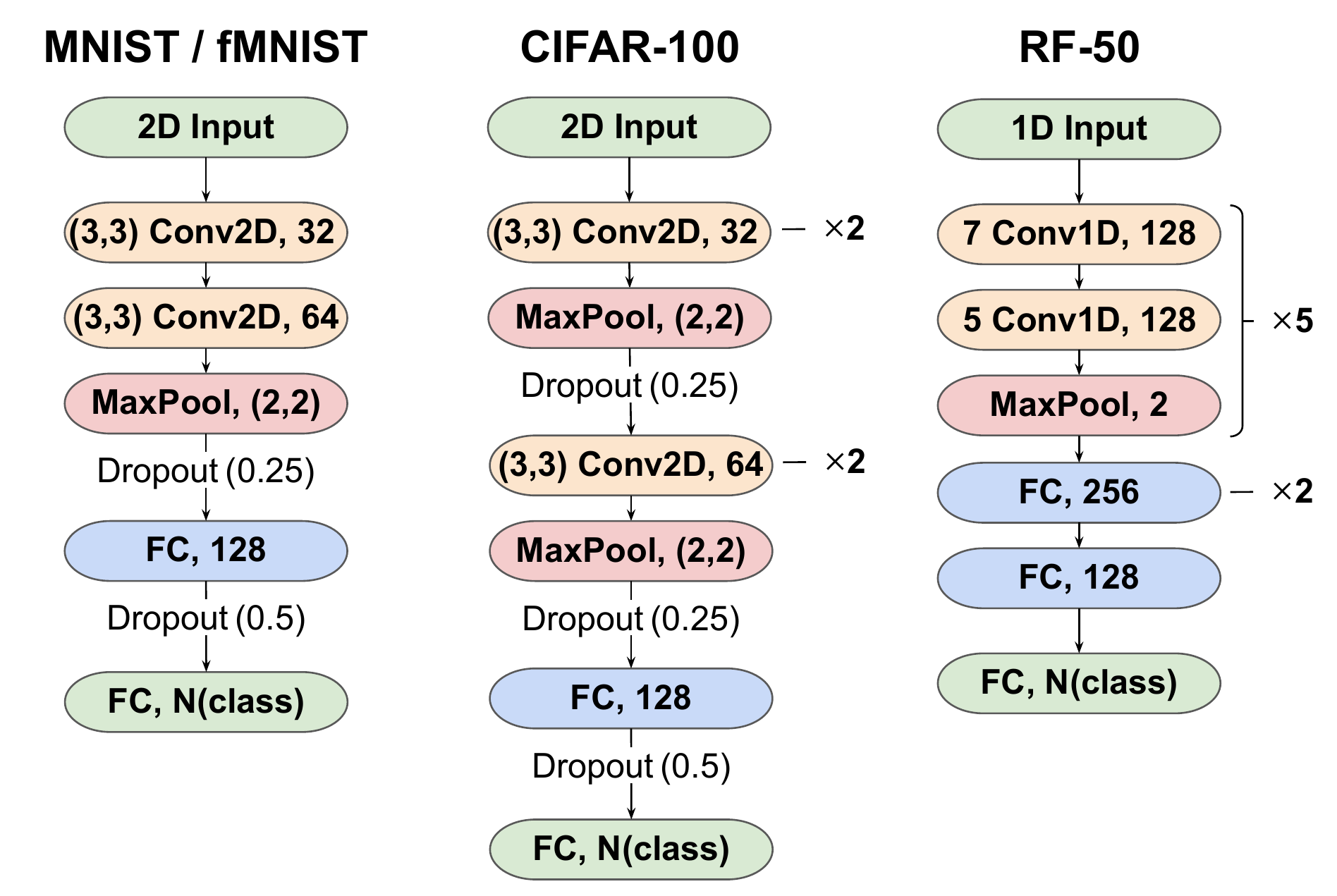}
    \vspace{-5mm}
    \caption{Network architectures used for MNIST/fashion MNIST, CIFAR-100, and RF-50 datasets.}
    \label{fig:baseline_models}
\end{figure}

\subsection{Implementation Details}
\label{sec:implementation}
For both MNIST, Fashion MNIST, and CIFAR-100 datasets, we employ convolutional neural networks suggested in the official Keras Examples Directory for the corresponding datasets~\cite{keras-examples}.
For the RF-50 dataset, we applied the baseline convolutional neural network proposed in~\cite{jian2020rfmls}. 
All three models are illustrated on~\Cref{fig:baseline_models}.

\begin{table}[]
    \centering
    \caption{Number of epochs for each dataset at different stages. We use the same numbers for MNIST and Fashion-MNIST.}
    \vspace{-2mm}
    \setlength{\tabcolsep}{4pt}
    \begin{tabular}{||c|c|c|c||}
    \hline
         Dataset & Pretraining & CD-KNet & Network Expansion \\
         \hline
         \hline
        (Fashion) MNIST & 50 & 20 & 30 \\
        \hline
        CIFAR-100 & 70 & 20 & 30 \\
        \hline
        RF-50 & 25 & 10 & 25 \\
        \hline
    \end{tabular}
    \label{tab:epochs}
    \vspace{-5mm}
\end{table}

We determined all the hyperparameters on a validation set via holding out $10\%$ of the training data. For all experiments, we use the same learning rates for Stages one and two. For Stage three, the network expansion stage, we use the same learning rate again to train the expanded last two layers and $0.1$ of the learning rate to fine tune the rest of the network. The proportion of $\mathcal{D}_l$ samples used in training network expansion is $20\%$. For MNIST and Fashion-MNIST, we use Adam \cite{DBLP:journals/corr/KingmaB14} optimizer with a learning rate of $0.01$. For CIFAR-100, we use Adam optimizer with a learning rate of $0.0001$. For RF-50, we followed the suggested hyperparameters by Jian et al.~\cite{jian2020rfmls}. The training epochs for Stage one  (pretraining), Stage two (CD-KNet), and Stage three (network expansion) are summarized in Table \ref{tab:epochs}. We set $\lambda = 10$ and the subsampling rates for MNIST, Fashion-MNIST, CIFAR-100 and RF-50 as $3\%$, $5\%$, $5\%$, $5\%$, respectively. 
%
We implement our method using Keras 2.2.4 \cite{chollet2015keras} and TensorFlow 1.14.0 \cite{tensorflow2015-whitepaper}.  
Our source code is publicly available.\footnote{ \texttt{\url{https://github.com/neu-spiral/OpenWorldKNet}}}

\subsection{Evaluation Metrics}
\label{sec:metrics}

We assess the quality of the clusters discovered by our proposed class discovery approach, CD-KNet-Exp, and the other competing methods described in~\Cref{sec:methods} on the unlabeled data $D_u$ using clustering evaluation metrics.  In particular, we report the following three clustering external evaluation metrics: clustering accuracy (ACC), normalized mutual information (NMI) and adjusted Rand index (ARI). These external evaluation criteria measure how well the discovered clusters match the ground truth unseen class labels. 
ACC is the standard clustering accuracy measure. NMI is the normalized mutual information between the estimated cluster pseudo-labels $\hat{y}_u$ and the ground-truth labels $y_u$: $\mathtt{NMI}(\hat{y}_u,y_u) = \frac{I(\hat{y}_u,y_u)}{\sqrt{H(\hat{y}_u)H(y_u)}}$, where $I(a,b)$ is the mutual information between $a$ and $b$, and $H(a)$ and $H(b)$ are the entropies of $a$ and $b$ respectively. Both $I$ and $H$ here are computed from empirical distributions. Finally, ARI is the adjusted Rand index, which measures the amount of overlap between two clustering solutions as follows:
%
    \begin{equation}
        \mathtt{ARI} = \frac{\mathtt{RI} - E[\mathtt{RI}]}{\max(\mathtt{RI}) - E[\mathtt{RI}]},\text{ where }\mathtt{RI} = \frac{a+b}{C^n_2},\text{ and}
    \end{equation}
    \noindent where $a$ is the number of pairs of data samples that belong to the same class w.r.t. both the ground-truth and pseudo-labels, $b$ is the number of data samples that belong to the different classes w.r.t. both the ground-truth and pseudo-labels, and $C^n_2$ is the total number of data sample pairs in the dataset.
All three metrics have values between zero and one.  The higher the value, the better the clustering agreement with the true labels.

\section{Results and Discussion}
\label{Sec:results and disscusions}
\subsection{Results on Benchmark Datasets and Radio Dataset}

\begin{table*}[htbp]
    \caption{Class discovery performance on all datasets.}
    \vspace{-2mm}
    \setlength{\tabcolsep}{5pt}
    \centering
     \begin{tabular}{||c | c c c | c c c | c c c | c c c ||} 
     \hline
     Datasets & \multicolumn{3}{|c|}{MNIST} & \multicolumn{3}{|c|}{Fashion-MNIST} & \multicolumn{3}{|c|}{CIFAR-100} &  \multicolumn{3}{|c||}{RF-50} \\
     \hline
     Methods\textbackslash Metrics & ACC & NMI & ARI & ACC & NMI & ARI & ACC & NMI & ARI & ACC & NMI & ARI \\
     \hline
     \hline
    SSCD & 0.840 & 0.629 & 0.649 & 0.552 & 0.386 & 0.316 & 0.201 & 0.193 & 0.064 & 0.524 & 0.194 & 0.184 \\
     \hline
    SSCD-Exp & 0.887 & 0.791 & 0.812 & 0.596 & 0.477 & 0.376 & 0.213 & 0.204 & 0.066 & 0.578 & 0.230 & 0.220 \\
     \hline
     \textcolor{red}{UCD-KNet} & 0.849 & 0.645 & 0.664 & 0.587 & 0.476 & 0.403 & 0.220 & 0.216 & 0.078 & 0.553 & 0.244 & 0.273 \\
     \hline
     \textcolor{red}{UCD-KNet-Exp} & 0.926 & 0.794 & 0.829 & 0.622 & 0.518 & 0.470 & 0.247 & 0.237 & 0.091 & 0.585 & 0.279 & 0.309 \\ 
     \hline
     \textcolor{blue}{CD-KNet} & 0.869 & 0.683 & 0.707 & 0.655 & 0.535 & 0.463 & 0.232 & 0.228 & 0.080 & 0.587 & 0.453 & 0.456 \\ 
     \hline
     \textcolor{blue}{CD-KNet-Exp} & \textbf{0.945} & \textbf{0.856} & \textbf{0.869} & \textbf{0.679} & \textbf{0.603} & \textbf{0.529} & \textbf{0.269} & \textbf{0.256} & \textbf{0.102} & \textbf{0.610} & \textbf{0.481} & \textbf{0.485} \\
     \hline
    \end{tabular}
    \label{tab:clusterperf}
    \vspace{-5mm}
\end{table*}

\noindent\textbf{Class Discovery Performance.}
In Table \ref{tab:clusterperf} we report the class discovery performance, in terms of ACC, NMI and ARI, of our proposed CD-KNet-Exp method, along with all other competing methods on all datasets. \emph{Note that in all datasets, CD-KNet-Exp performs the best against all methods w.r.t.~all three clustering metrics.}

From this table, we can also observe the effect of network expansion compared to versions without (`Exp').  
Note that all the methods which incorporate network expansion gain performance increase over their counterpart, showing the effectiveness of this strategy. 

Another interesting observation from Table \ref{tab:clusterperf} is that our CD-KNet-Exp and UCD-KNet-Exp achieve the 1st and 2nd best performance consistently on all datasets.  This demonstrates that (a) only using the unsupervised part in our HSIC objective already leads to better separable latent feature embeddings than other methods; and (b) introducing supervised knowledge from old classes further improves the performance of only using unsupervised information. 



\begin{table*}[htbp]
    \caption{Comparing with full unsupervised learning methods.}
    \label{tab:unsup}
    \vspace{-2mm}
    \centering
     \begin{tabular}{||c | c c c | c c c | c c c ||} 
     \hline
     Datasets & \multicolumn{3}{|c|}{MNIST} & \multicolumn{3}{|c|}{Fashion-MNIST} & \multicolumn{3}{|c|}{CIFAR-100} \\
     \hline
     Methods\textbackslash Metrics & ACC & NMI & ARI & ACC & NMI & ARI & ACC & NMI & ARI \\
     \hline\hline
    DEC & 0.873 & 0.835 & 0.822 & 0.620 & 0.566 & 0.522 & 0.205 & 0.166 & 0.073  \\
    \hline
    DAC & 0.894 & 0.843 & 0.865 & 0.642 & 0.588 & 0.543 & 0.247 & 0.195 & 0.088  \\
     \hline
     \textcolor{blue}{CD-KNet-Exp} & \textbf{0.945} & \textbf{0.856} & \textbf{0.869} & \textbf{0.679} & \textbf{0.603} & \textbf{0.529} & \textbf{0.269} & \textbf{0.256} & \textbf{0.102} \\
     \hline
    \end{tabular}
    \vspace{-3mm}
\end{table*}

\noindent\textbf{Comparing with Unsupervised Deep Learning Methods.}
In order to show that our method provides an effective way of incorporating knowledge learned from old classes, we compare our method with two state-of-the-art deep learning based unsupervised clustering methods, DEC \cite{xie2016unsupervised} and DAC \cite{chang2017deep}, on three benchmark datasets in Table~\ref{tab:unsup}. To mimic our problem setting, we pretrain the deep neural networks used in DEC and DAC with \emph{old} classes, and then we perform these algorithms on \emph{new} classes. Table~\ref{tab:unsup} shows that \emph{CD-KNet-Exp beats both of the methods on all three benchmarks}. Although we enforce DEC and DAC to incorporate knowledge from old classes by pretraining, our method still outperforms both DEC and DAC. 

\begin{figure*}[ht]
\begin{subfigure}{0.48\textwidth}
  \centering
  \includegraphics[width=.8\linewidth]{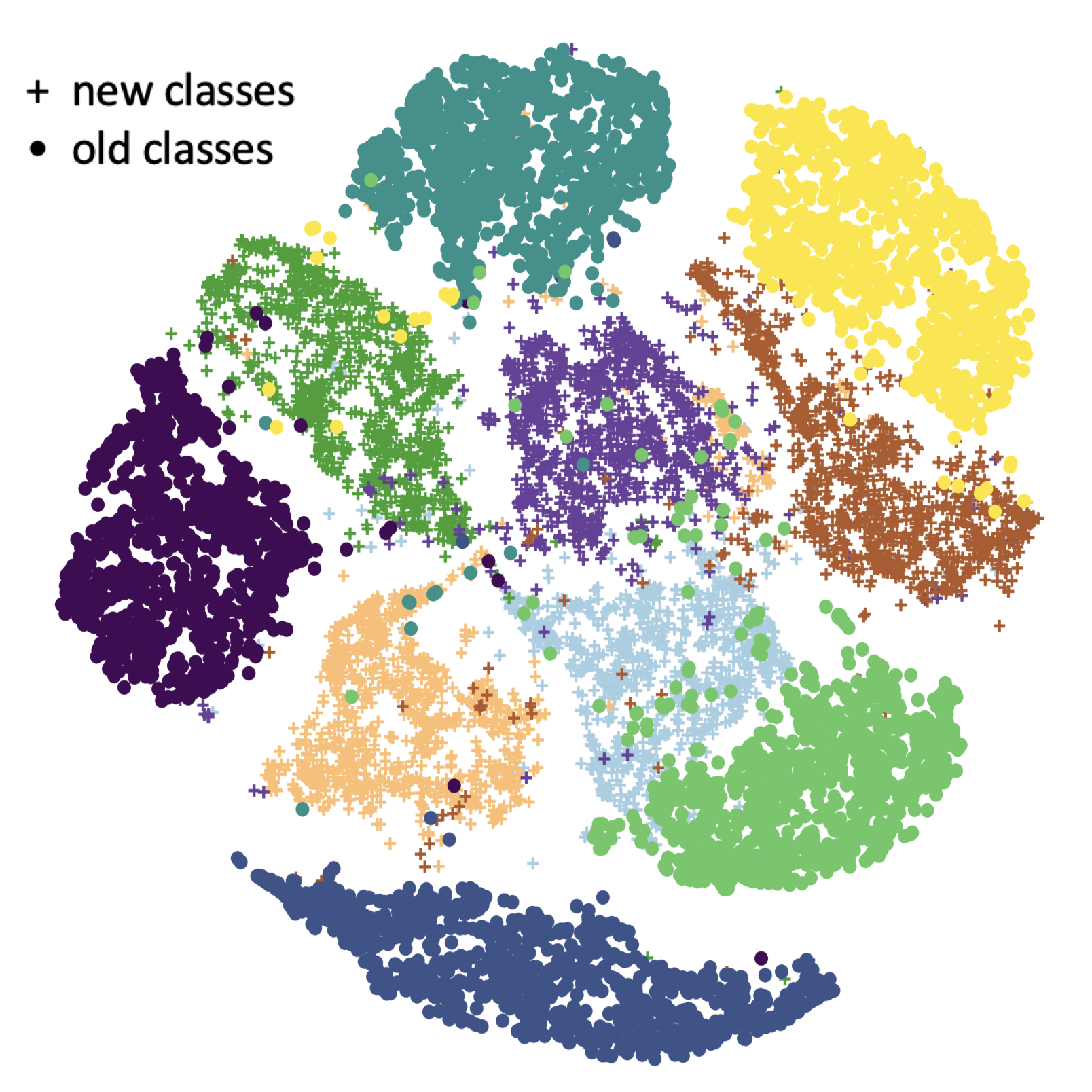}  
  \caption{CD-KNet-Exp}
  \label{fig:sub-first}
\end{subfigure}
\hfill
\begin{subfigure}{0.48\textwidth}
  \centering
  \includegraphics[width=.8\linewidth]{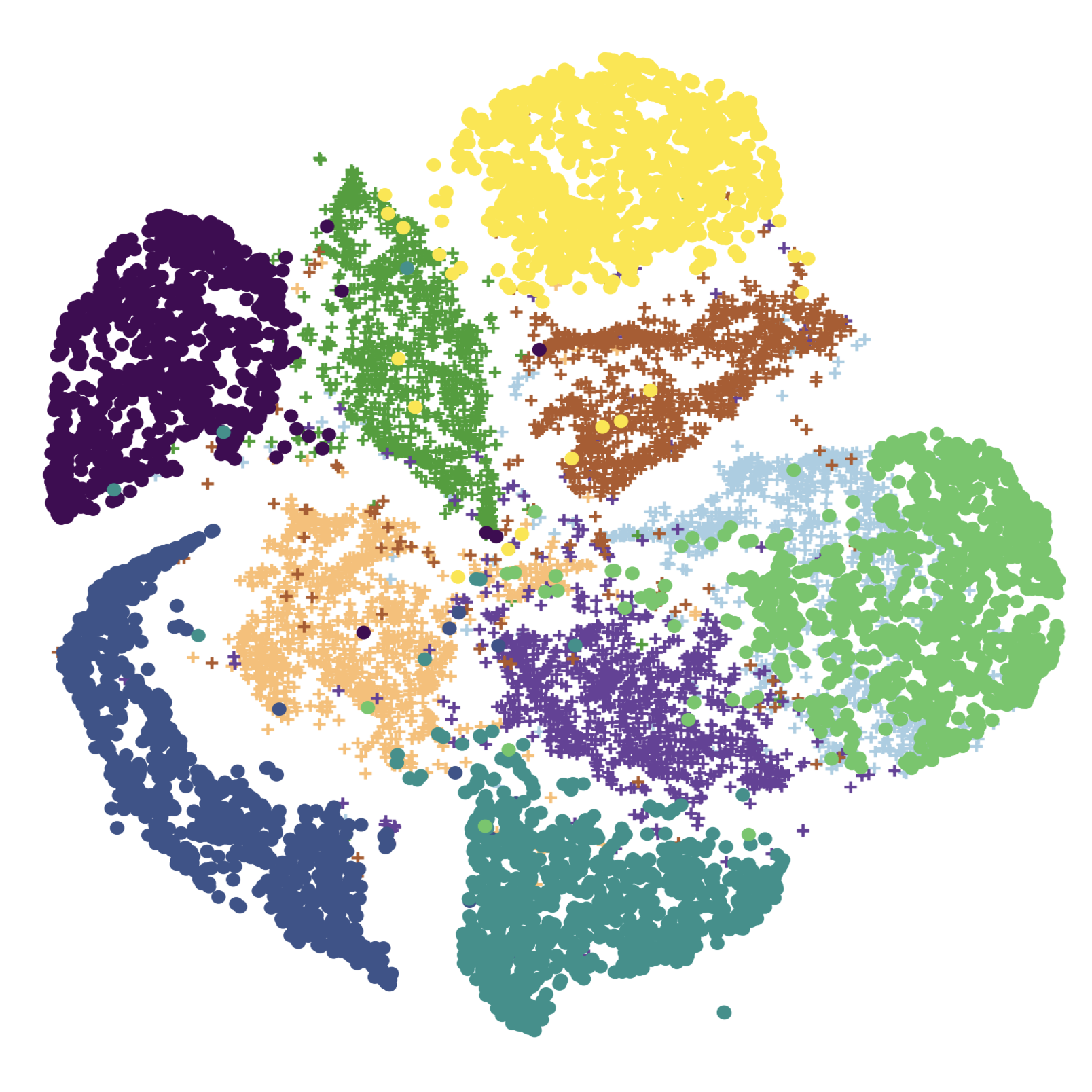}  
  \caption{SSCD-Exp}
  \label{fig:sub-second}
\end{subfigure}
\vspace{-1mm}
\caption{t-SNE visualization of latent embeddings of CD-KNet-Exp and SSCD-Exp. CD-KNet-Exp 1) produces better separable latent embeddings than SSCD-Exp; 2) has clearer boundaries between old and new classes.}
\label{fig:tsne}
\vspace{-3mm}
\end{figure*}

\noindent \textbf{Visualization of Latent Embeddings.}
In Figure \ref{fig:tsne}, we investigate the latent embeddings learned by our method CD-KNet-Exp against the state-of-the-art method for class discovery SSCD-Exp on the MNIST dataset.  We plot the embeddings using t-SNE \cite{maaten2008visualizing} visualization of the latent feature embedding after expansion for \emph{all} classes, where `$+$' represents new classes, `$\bullet$' represents old classes, and different colors represent different classes. Observe that CD-KNet-Exp separates the classes better than SSCD-Exp. Also, we notice that SSCD mixes one of the new classes (light blue) and old classes (green) together. This explains why there is an accuracy drop on the old classes (which we will explain later in Table~\ref{tab:extension}). Recall that SSCD directly uses feature extractor learned on old classes, so that it has an underlying drawback of mistaking new classes to old ones when their samples are similar in the latent embedding space. In contrast, CD-KNet-Exp takes both labeled old classes and unlabeled new classes into consideration, pulling them apart in the learned latent embedding space by optimizing the CD-KNet HSIC based objective.

\subsection{Influence of Controllable Parameters.}
Our method, CD-KNet-Exp has several controllable parameters: the control parameter $\lambda$ between supervised and unsupervised HSIC, network expansion, and the subsampling factor.  In this subsection, 
explore how these parameters affect the final class discovery performance.

\begin{table}[t]
    \caption{Influence of $\lambda$ on Fashion MNIST dataset.}
    \label{tab:lambda}
    \vspace{-2mm}
    \setlength{\tabcolsep}{12pt}
    \centering
     \begin{tabular}{||c|c|c|c||} 
     \hline
     Metric\textbackslash$\lambda$ & 10  & 0 & $\infty$\\ 
     \hline\hline
     ACC & 0.655 & 0.587 & 0.544 \\
     \hline
     NMI & 0.535 & 0.476 & 0.375 \\
     \hline 
     ARI & 0.463 & 0.403 & 0.276 \\
     \hline
    \end{tabular}
    \vspace{-5mm}
\end{table}

\noindent\textbf{Influence of $\lambda$.}
In Equation~(\ref{eq:ow-hsic}), we have a balance factor $\lambda$ to control how much weight to put on the supervised part of our HSIC based objective. In practice, we find that the algorithm is not sensitive to the value of $\lambda$ for a wide range ($\lambda \in [1, 100]$) of values, so we just set $\lambda=10$ as a representative value in our experiments. However, we observe clear differences in performance when there is only unsupervised information ($\lambda = 0$), supervised and unsupervised information ($\lambda = 10$) and only supervised information ($\lambda = \infty$, in practice, we enforce this by  removing the unsupervised term). 

Table~\ref{tab:lambda} reports the class discovery performance in terms of ACC, NMI and ARI of CD-KNet-Exp for $\lambda=10$, $\lambda=0$, and $\lambda=\infty$ on Fashion-MNIST dataset. When Comparing $\lambda=10$ and $\lambda=0$, we are actually comparing CD-KNet-Exp and UCD-KNet-Exp. The better performance of CD-KNet-Exp over unsupervised UCD-KNet-Exp showcases the importance of supervision in our HSIC based objective, which is also demonstrated in Table~\ref{tab:clusterperf}. Interestingly, $\lambda=\infty$ results in the worst performance, indicating that  only using supervised HSIC to tune the network tends to overfit 
to old classes, resulting in poor performance on the new classes. 

\noindent\textbf{Influence of Network Expansion.}
We have already shown how network expansion helps new class discovery in Table~\ref{tab:clusterperf}. 
In this subsection, we investigate the effect of expansion 
on the classification accuracy on the previous old classes. 
Table~\ref{tab:extension} shows the cluster accuracy and classification accuracy of new and old classes respectively, before expansion and after expansion on the MNIST dataset. Both CD-KNet-Exp and UCD-KNet-Exp incorporate network expansion strategy well with clustering accuracy increase on new classes and almost no classification accuracy decrease on old classes. In contrast, although SSCD achieves better cluster accuracy on new classes after expansion, the classification accuracy of SSCD on old classes drops by about $10\%$. 


\begin{figure}[htbp]
    \centering
    \includegraphics[width=0.7\linewidth]{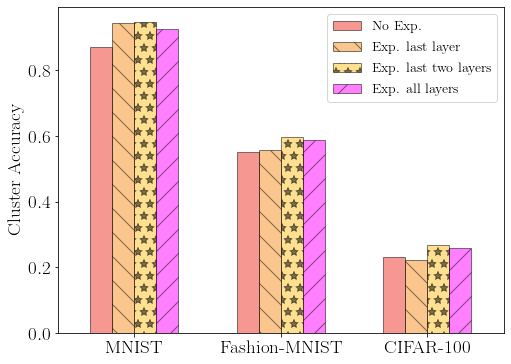}
    \vspace{-2mm}
    \caption{Comparison of different expansion strategies and no expansion. Expansion of the last two layers of the network performs the best among all strategies.}
    \label{fig:expansion_result}
    \vspace{-5mm}
\end{figure}

\noindent \textbf{Comparing Different Expansion Strategies.} To justify the reason why we choose to expand the last two layers of the DNN, we compare 3 different expansion strategies and completely no expansion. The 3 expansion strategies are: (a) Expand only the last layer; (b) Expand the last two layers (ours); and, (c) Expand all layers in the DNN (for convolutional layers, we double its number of filters). Figure~\ref{fig:expansion_result} demonstrates that our strategy, expanding the last two layers, indeed performs the best. Moreover, note that an all expansion strategy performs better than no expansion. On MNIST dataset, expanding only the last layer is comparable with our strategy. However, on Fashion-MNIST and CIFAR-100, there is a performance gap between expanding only the last layer and the last two layers. This observation suggests that the feature extractor has enough representation capacity for relatively simple datasets like MNIST, but for complex datasets, we need to expand the feature extractor to allocate more representation power for new classes. On the other hand, expanding all layers which increases the representation capacity of the DNN leads to worse performance than just expanding the last two layers. This is a case of overfitting, which may harm the performance 
when we expand the DNN too much.

\begin{figure*}[htbp]
  \centering
  \includegraphics[width=1.0\linewidth]{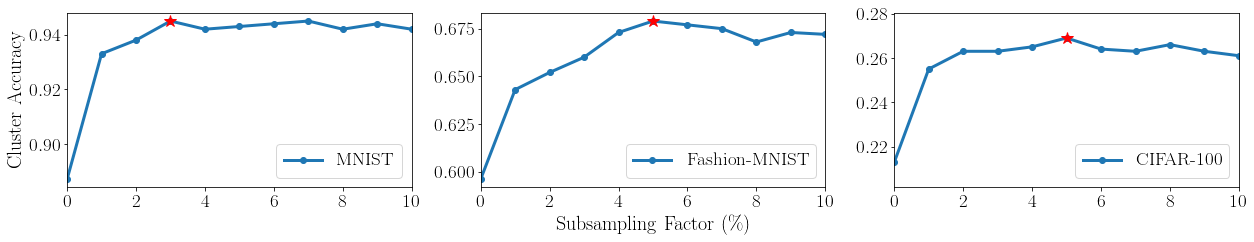}
  \vspace{-6mm}
  \caption{Influence of subsampling factor w.r.t. $\mathtt{ACC}$ measure. CD-KNet-Exp is able to perform well using only 3\%, 5\%, 5\% of the MNIST, Fashion-MNIST and CIFAR-100 datasets, respectively.}
  \label{fig:subsampling}
  \vspace{-4mm}
\end{figure*}

\noindent \textbf{Influence of Subsampling Factor.}
As mentioned in Section \ref{subsec:solve_hsic}, calculating the kernel matrix for all the data is computationally expensive, so we adopt a subsampling strategy~\cite{wu2020deep} to reduce the computation. In Fig \ref{fig:subsampling}, we explore how the subsampling factor will affect the final performance. For each benchmark dataset, we set the subsampling factor from $0\%$ (equivalent to SSCD) to $10\%$. The trends are very similar for all datasets: There is a notable jump in accuracy from $0\%$ to $1\%$ and then the accuracy increase slows down until it reaches the best performance ($3\%$ for MNIST and $5\%$ for Fashion-MNIST and CIFAR-100). After that, the accuracy tends to be stable. These empirical results show that CD-KNet only needs a small portion of the data to capture the whole picture, generalizing well to the entire dataset.

\begin{table}[t]
    \caption{Influence of Network Expansion on MNIST dataset w.r.t. $\mathtt{ACC}$ measure}
    \label{tab:extension}
    \vspace{-2mm}
    \centering
     \begin{tabular}{||c|c|c|c|c||} 
     \hline
      & \multicolumn{2}{|c|}{New Classes}  & \multicolumn{2}{|c||}{Old Classes} \\ 
     \hline
     Method\textbackslash Stage &  Before &  After &  Before &  After \\
     \hline\hline
    SSCD-Exp & 0.840 & 0.887 & 0.988 & 0.892 \\
     \hline
    U-KNet-Exp & 0.849 & 0.926 & 0.988 & 0.983 \\
     \hline
     CD-KNet-Exp  & 0.869 & 0.945 & 0.988 & 0.983 \\
     \hline
    \end{tabular}
    \vspace{-4mm}
\end{table}




\section{Conclusions}\label{Sec:conclusions}
In this paper, we introduced CD-KNet-Exp for addressing the open-world class discovery problem.  Empirical results on MNIST, Fashion-MNIST, CIFAR-100 and a real-world radio frequency fingerprinting dataset, RF-50, show that CD-KNet-Exp can discover new classes with clustering performance better than all competing methods.
In particular, it outperforms competitors  over  three  benchmark datasets  by  a $2\%$-$12\%$ margin.


\section{Acknowledgements}
The authors gratefully acknowledge support by the National Science Foundation (grant CCF-1937500).

\bibliographystyle{IEEEtran}
\bibliography{IEEEabrv,ref}{}
\vspace{12pt}

\end{document}